\documentclass[10pt,twocolumn,letterpaper]{article}

\usepackage{cvpr}
\usepackage{times}
\usepackage{epsfig}
\usepackage{graphicx}
\usepackage{amsmath}
\usepackage{amssymb}

\usepackage[pagebackref=true,breaklinks=true,letterpaper=true,colorlinks,bookmarks=false]{hyperref}

\cvprfinalcopy % *** Uncomment this line for the final submission

\def\cvprPaperID{****} % *** Enter the CVPR Paper ID here

\usepackage{kotex}

\begin{document}

%%%%%%%%% TITLE
\title{Coloring With Limited Data: \\
Few-Shot Colorization via Memory-Augmented Networks}

% \author{
% Seungjoo Yoo\textsuperscript{1}\\
% \AND
% Junsoo Lee\textsuperscript{1}\\
% \and
% Hyojin Bahng\textsuperscript{1}\\
% \AND
% Jaehyuk Chang\textsuperscript{2}\\
% \and
% Sunghyo Chung\textsuperscript{1}\\
% \AND
% Jaegul Choo\textsuperscript{1}\\
% \and
% \textsuperscript{1} Korea University
% \and
% \textsuperscript{2} NAVER WEBTOON Corp.\\
% %\tt\small{{seungjooyoo, hjj552, s94021, junsulee, jchoo}@ korea.ac.kr, jaehyuk.chang@webtoonscorp.com}
% }

% \author{\spacedlowsmallcaps{Seungjoo Yoo\textsuperscript{1}\hspace{1.08cm}Hyojin Bahng\textsuperscript{1}\hspace{1cm}Sunghyo Chung\textsuperscript{1}} \\
% \vspace{5mm}
% \spacedlowsmallcaps{Junsoo Lee\textsuperscript{1}\hspace{1.2cm}Jaehyuk Chang\textsuperscript{2}\hspace{1.2cm}Jaegul Choo\textsuperscript{1}} \\
% \spacedlowsmallcaps{\textsuperscript{1} Korea University\hspace{0.8cm}\textsuperscript{2} NAVERWEBTOON Corp.}}

\author{Seungjoo Yoo\textsuperscript{1}, Hyojin Bahng\textsuperscript{1}, Sunghyo Chung\textsuperscript{1}, Junsoo Lee\textsuperscript{1}, Jaehyuk Chang\textsuperscript{2}, Jaegul Choo\textsuperscript{1}\\
\\
\textsuperscript{1} Korea University\hspace{0.8cm}\textsuperscript{2} NAVERWEBTOON Corp.}

\maketitle

%%%%%%%%% ABSTRACT 
\begin{abstract} 
Despite recent advancements in deep learning-based automatic colorization, they are still limited when it comes to few-shot learning. Existing models require a significant amount of training data. To tackle this issue, we present a novel memory-augmented colorization model \textbf{MemoPainter} that can produce high-quality colorization with limited data. In particular, our model is able to capture rare instances and successfully colorize them. We also propose a novel threshold triplet loss that enables unsupervised training of memory networks without the need of class labels. Experiments show that our model has superior quality in both few-shot and one-shot colorization tasks. 
%   Despite recent advancements in deep learning-based automatic colorization, they are still limited when it comes to few-shot learning. Existing models require a significant amount of training data to produce high-quality results. To tackle this issue, we present a novel memory-augmented colorization model \textbf{MemoPainter} that can produce high-quality colorization even with limited data. Our model is able to capture rare instances and successfully colorize them. Also, we propose a novel Threshold Triplet Loss that enables unsupervised training of memory networks without the need for class labels. Experiments show that our model has superior quality in both few-shot and one-shot colorization tasks.
%   The contributions of \textit{MemoPainter} are (i) it works well even in one/few-shot settings, (ii) is fully automatic and does not rely on user hints, and (iii) introduces a novel triplet loss that allows unsupervised training of memory networks, which normally require supervised labels. 
\end{abstract}
% small figure
%\begin{figure}[t]
%\begin{center}
%   %\includegraphics[width=0.8\linewidth]{egfigure.eps}
%\end{center}
%   \caption{first figure here}
%\label{fig:long}
%\label{fig:onecol}
%\end{figure}
%%%%%%%%% BODY TEXT
\section{Introduction}

When Dorothy stepped into Land of Oz in the 1939 movie \textit{Wizard of Oz}, a transition from black and white to vibrant colors makes it one of the most breathtaking moments in the history of cinema. There is no doubt to colors being an effective tool of expression, but they usually come at a cost. Coloring images is one of the most laborious and expensive stages when making modern day animation movies and comics. Automating the colorization process can help to reduce both cost and time required in producing comics or animated movies.

Despite advances in deep learning-based colorization models~\cite{zhang2016colorful,guadarrama2017pixcolor,isola2017image,zhang2017real}, they are still limited when it comes to real-world applications like coloring animations and cartoons. There exist two main problems that make it difficult to use deep colorization models in real-world settings. 
% Emphasize "few-shot"

% teaser figure
\begin{figure}[t]
\begin{center}
\includegraphics[width=0.9\linewidth]{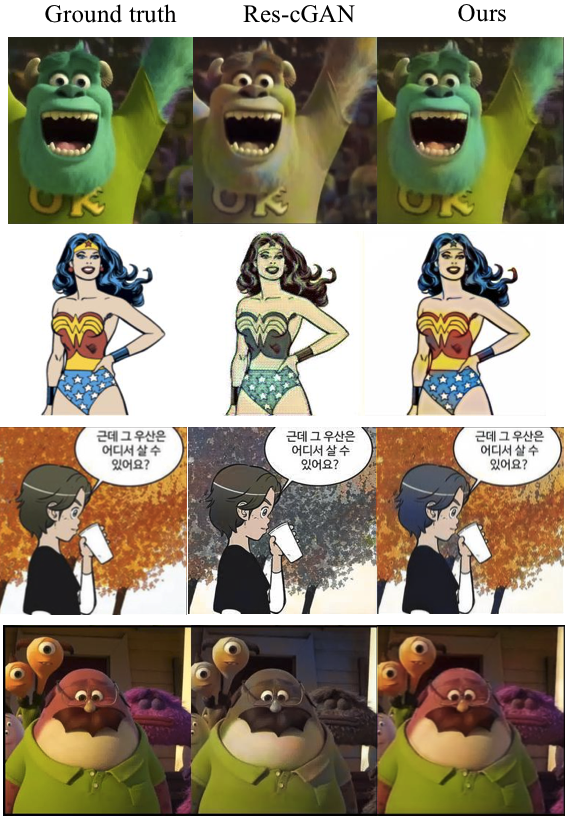}
\end{center}
   \caption{Not much data in your hands? Make the most out of your limited data with our fully-automatic colorization model \textit{MemoPainter.} (Res-cGAN is \textit{MemoPainter} without our proposed memory networks.)}
   \vspace{-0.03em}
\label{fig:long}
\label{fig:onecol} \label{fig:first_figure}
\end{figure}

First, data for animations and cartoons are often limited, but training deep learning-based colorization models requires a large amount of data. Cartoon images are difficult to create because they must be drawn and intricately colored by hand. In contrast, obtaining real-world images is easier because they can be taken by a camera and simply converted to grayscale. This leads to cartoon data not being as abundant as real-world images. Numerous existing colorization models are trained on real-world images, and their application is mostly limited to coloring old legacy photographs. This task is no longer needed because modern-day photographs are produced in color. Thus, learning to color animations and cartoons with little data would allow a more practical application of deep colorization models. 
% For identity-preserving colorization tasks, acquiring data is expensive and time-consuming as sufficient data need to be collected for each identity. In particular, generating sufficient data for animations and cartoons is highly expensive as they need to be produced by professional animators, and different training data is needed for different animation style and content. In this paper, we present a few-shot colorization model that make the most out of limited data.

\begin{figure}[t]
\begin{center}
\includegraphics[width=1.0\linewidth]{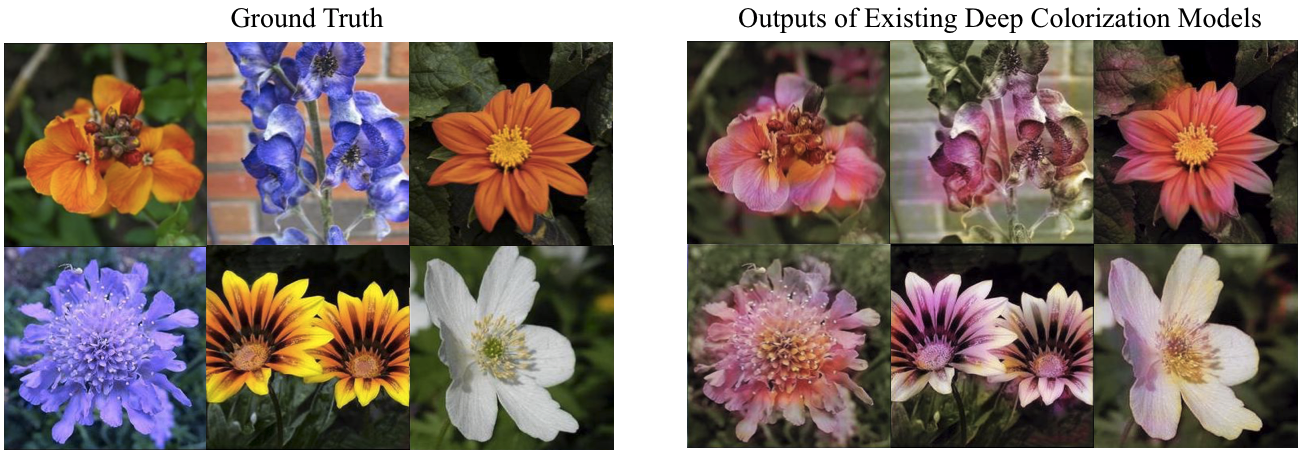}
\end{center}
   \caption{\textbf{Dominant color effect commonly encountered by deep colorization models.} Deep colorization models tend to ignore diverse colors present in a training set and opt to learn only a few dominant colors. Using the most dominant color can be effective in minimizing the overall loss but yields unsatisfactory results. One can see that the outputs of~\cite{zhang2017real} are dominated by the most prevalent color (red).}
\label{fig:dominant_color}
\end{figure}

% Emphasize "rare instance"
Second, existing colorization models ignore rare instances present in data and opt to learn the most frequent colors to generalize over the data. Remembering rare instances is important when diverse characters appear in a story that we want to color. Rare side characters will be ignored by colorization networks and all side characters will be colored similarly to the main character. Existing colorization models suffer from the dominant color effect, illustrated in Fig.~\ref{fig:dominant_color}. This effect occurs when a colorization model only learns to color with a few dominant colors present in the training set. This leads to existing models being unable to preserve \textit{color identity}, which we define as the distinctive colors that separate a particular object class from the other. An example of color identity can be found in flowers. Different flower classes are distinguished by both their color and shape (buttercups are yellow and roses are red). Coloring in the most dominant color may succeed in producing plausible and natural-looking outputs, but each image loses its color identity. 
% small figure: dominant color effect

We aim to alleviate these problems with our novel memory-augmented colorization model \textit{MemoPainter}. To the best of our knowledge, there has been no colorization networks augmented by external neural memory networks. The main contributions of this paper include:

(1)	Our model can learn to color with little data, allowing one-shot or few-shot colorization. This is possible because our memory networks extract and store useful color information from a given training data. When an input is given to our model, we can query our external memory networks to extract color information relevant to coloring the input. 

% Performance on rare instances
(2)	Our model can capture images of rare classes and suffer less from the dominant color effect, which previous methods have not been able to accomplish. 

(3)	We present a novel threshold triplet loss, which allows training of memory networks in an \textit{unsupervised} setting. We do not need labeled data for our model to successfully colorize images. 

% (small) figure showing cartoon abstraction
%\begin{figure}[t]
%\begin{center}
%\includegraphics[width=0.7\linewidth]{cvpr2019AuthorKit2/latex/i%mages/comic_simple.png}
%\end{center}
%   \caption{Cartoons(right) are extremely simplified, which %makes it more difficult for neural networks to learn meaningful %information from them when compared to real-world images(left)}
%\label{fig:long}
%\label{fig:onecol}
%\end{figure}

%long figure
\begin{figure*}
\begin{center}
\includegraphics[width=1.0\linewidth]{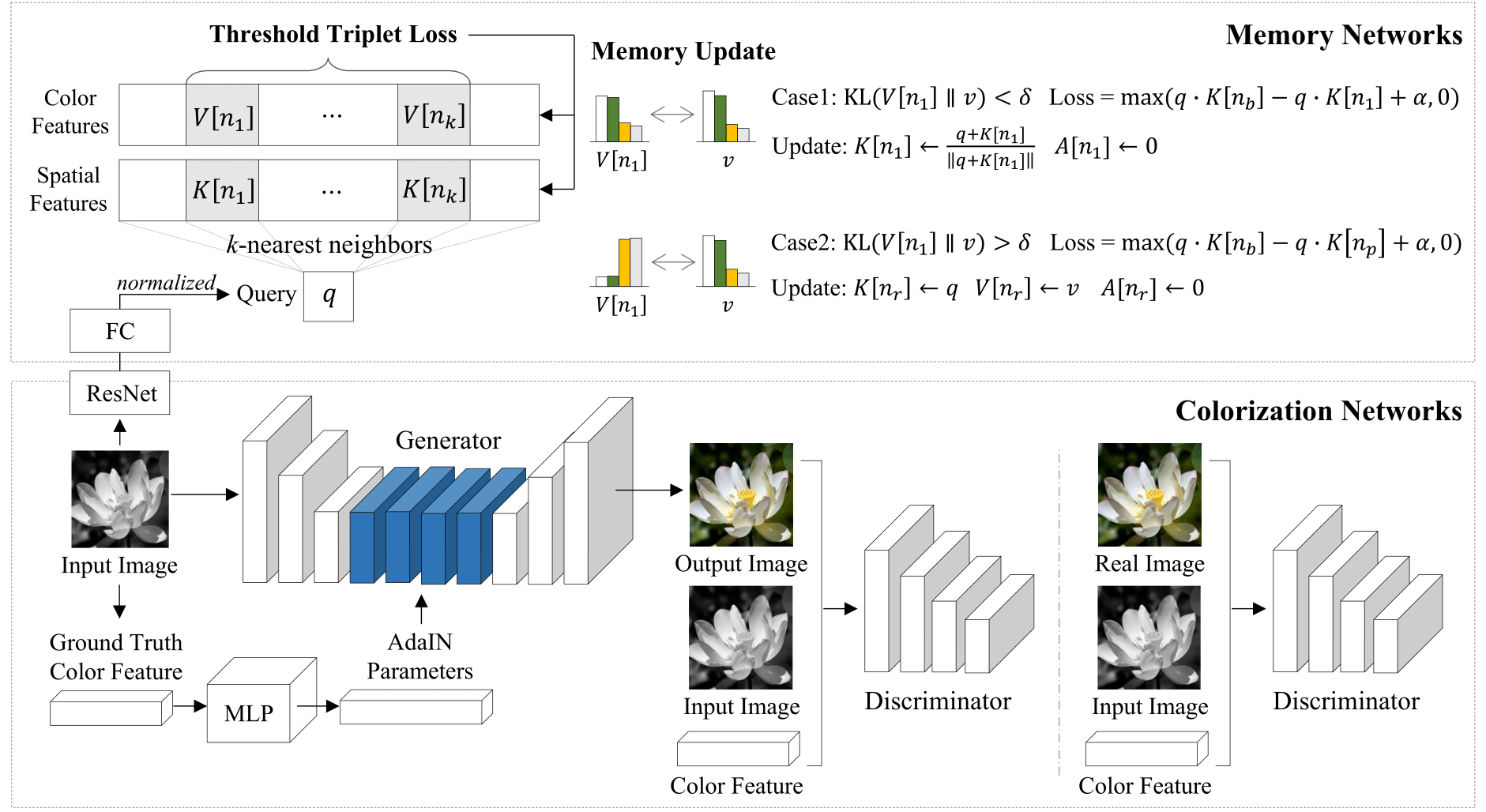}
\end{center}
   \caption{\textbf{Our proposed \textit{MemoPainter} model.} Our model consists of memory networks and colorization networks. During training, memory networks learn to retrieve a color feature that best matches the ground-truth color feature of the query image, while the colorization networks learn to effectively inject the color feature to the target grayscale image. During test time, we retrieve the top-1 color feature from our memory and give it as a condition to the trained generator.
   } 
\label{fig:full_model}
\end{figure*}

%------------------------------------------------------------------------
\section{Related Work}

\paragraph{Deep Learning-Based Colorization.}
Existing colorization methods~\cite{zhang2016colorful, zhang2017real, isola2017image} use deep neural networks to improve colorization performance. Zhang \textit{et al.}~\cite{zhang2016colorful} train convolutional neural networks and re-weights the loss function at training time to emphasize rare colors, yielding more vibrant results. Zhang \textit{et al.}~\cite{zhang2017real} incorporate local and global color hint information to increase colorization performance, which enables interactive colorization during test time. Isola \textit{et al.}~\cite{isola2017image} use conditional generative adversarial networks to improve colorization performance as well as other image-to-image translation tasks. Although existing deep colorization methods produce high-quality results, they inevitably require large-scale data to train the deep neural networks. However, preparing abundant training data for real-world applications such as animation colorization is highly expensive as they need to be produced by professional animators. Moreover, deep colorization networks are successful \textit{on average}(i.e., successful in coloring prominent objects yet failing in coloring rare instances). No previous studies have tackled few-shot colorization on rare instances, which is the main focus of this work.

\paragraph{Memory Networks.}
Several approaches have augmented neural networks with an external memory module to store critical information over long periods of time. It has been applied to solve algorithmic
problems~\cite{graves2014neural}, perform natural-language question answering~\cite{sukhbaatar2015end, kumar2016ask, miller2016key}, and allow life-long and one-shot learning, especially in remembering rare events~\cite{kaiser2017learning}.
Other approaches have applied memory networks to store image data, specifically for image captioning~\cite{park2017attend, park2018towards}, summarization~\cite{Kim:2019:NAACL-HLT}, image generation~\cite{kim2018memorization}, and video summarization~\cite{lee2018memory}. We are the first to augment colorization networks with memory networks to allow few-shot learning in image colorization.

\paragraph{Conditional Generative Adversarial Networks.} Generative adversarial networks (GANs)~\cite{goodfellow2014generative} have achieved remarkable success in image generation. The key to its success lies in its adversarial loss, where the discriminator tries to distinguish between real and fake images while the generator tries to fool the discriminator by producing realistic fake images. Several studies leverage \textit{conditional} GANs in order to generate samples conditioned on the class~\cite{mirza2014conditional, odena2016semi, odena2016conditional}, text description~\cite{reed2016generative, han2017stackgan, Tao18attngan}, domain information~\cite{StarGAN2018, pumarola2018ganimation}, input image~\cite{isola2017image,Cho2019Image}, or color features~\cite{bahng2018coloring}. In this paper, we adopt the adversarial loss conditioned on a grayscale image and its color feature extracted from our memory module to generate colored images indistinguishable from real images.

%------------------------------------------------------------------------
% 내용 추가 하기 노벨티 훨씬 강조하기
\section{Proposed Method}
As illustrated in Fig.~\ref{fig:full_model}, our model \textit{MemoPainter} is composed of two networks: memory networks and colorization networks. \textit{MemoPainter} is the first model to augment colorization networks with memory to remember rare instances and produce high-quality colorization with limited data. Our memory networks are distinguished from previous approaches by how its key and value memory are constructed. We also introduce a new threshold triplet loss (TTL), which allows unsupervised training of memory networks without additional class label information. Finally, our colorization networks utilize adaptive instance normalization (AdaIN)~\cite{huang2017arbitrary} to boost colorization performance.

%------------------------------------------------------------------------

\subsection{Memory Networks}
\label{memory_net}
 We construct memory networks to store three different types of information: key memory, value memory, and age. A key memory $K$ stores information about spatial features of input data. The key memory is used to compute the cosine similarity with input queries. A value memory $V$ stores color features which are later used as the condition for the colorization networks. Both memory components are extracted from the training data. An age vector $A$ keeps track of the age of items stored in memory without being used. Our entire memory structure $M$ can be denoted as
\begin{equation}
\begin{aligned}
M = ({ K }_{ 1 },{ V }_{ 1 },{ A }_{ 1 }),({ K }_{ 2 },{ V }_{ 2 },{ A }_{ 2 }),...,({ K }_{ m },{ V }_{ m },{ A }_{ m }),
\end{aligned}
\end{equation}
where $m$ represents the memory size. Our memory networks are inspired by the previously proposed architecture~\cite{kaiser2017learning}.

A query $q$ is constructed by first passing the input image $X$ through ResNet18-pool5 layers~\cite{he2016deep} pre-trained on ImageNet~\cite{deng2009imagenet}. It is denoted as ${ X }_{ rp5 }\in { \mathbb{R} }^{ 512 }$. We use feature vectors from pooling layers to summarize spatial information. For instance, a rose should be perceived as the same rose regardless of where it is spatially positioned in an image. We pass the feature representation through a linear layer with learnable parameters ${ W }\in { \mathbb{R} }^{ 512\times 512 }$ and  ${ b }\in { \mathbb{R} }^{ 512 }$. Finally, we normalize the vector to construct our query $q$ as 
\begin{equation}
\begin{aligned}
q = W{X}_{ rp5 }+b, \:
q = \frac { q }{ \left\| q \right\|  },
\end{aligned}
\end{equation}
where $\left\Vert q\right\Vert _{2}=1$. Given $q$, the memory networks compute the $k$ nearest neighbors with respect to cosine similarity between the query and the keys $d_{i}=q\cdot K\left[i\right]$, i.e., 
\begin{equation}
\begin{aligned}
\mathrm{NN}(q,M)=\mathrm{argmax_{\mathit{i}}\:\mathit{q\cdot K\left[i\right]}}, \\
(n_{1}, ..., n_{k})=\mathrm{NN}_{k}(q,M),
\end{aligned}
\end{equation}
and returns the nearest value $V[{n}_{1}]$, which is later used as the condition for the colorization networks.

% \paragraph{Key Memory.}
% Our key memory consists of feature maps from ResNet18~\cite{he2016deep} pre-trained on ImageNet~\cite{deng2009imagenet}. Each key memory slot is made up of feature representations of input image $X$ passed through ResNet18-pool5 layers. It is denoted as ${ X }^{ rp5 }\in { \mathbb{R} }^{ 512 }$. We use feature vectors from pooling layers to summarize spatial information. For instance, a rose should be perceived as the same rose regardless of where it is spatially positioned in an image. 
% % An $i$-th slot of the key memory ${K}_{i}$ is denoted as 
% % \begin{equation}
% % \begin{aligned}
% % { K }_{ i }={ X }_{ i }^{ rp5 }.
% % \end{aligned}
% % \end{equation}
% Our query $q$ is constructed by passing the feature representation through a linear layer 
% \begin{equation}
% \begin{aligned}
% { q=W }_{ i }{ X }_{ i }^{ rp5 }+{ b }_{ i },
% \end{aligned}
% \end{equation}
% where we introduce learnable parameters ${ W }_{ i }\in { \mathbb{R} }^{ 512\times 512 }$ and  ${ b }_{ i }\in { \mathbb{R} }^{ 512 }$.

\paragraph{Color Features.}
We leverage two variants to represent color information stored in value memory: color distributions and RGB color values. The former has the form of color distributions over 313 quantized color values, denoted as ${ C }_{ dist }\in { \mathbb{R} }^{ 313 }$. It is computed by converting an input RGB image to the CIE \emph{Lab} color space and quantizing the $ab$ values into 313 color bins. We use the previously proposed parametrization~\cite{zhang2016colorful} to quantize $ab$ values. Color distributions are suitable for images with diverse colors and intricate drawings.

The second variant we use is a set of ten dominant RGB color values of an image denoted as ${C}_{ RGB }\in { \mathbb{R} }^{ 10\times 3 }$, which is extracted from input images by utilizing a tool called Color Thief.~\footnote{http://lokeshdhakar.com/projects/color-thief/} 
Using ${ C }_{ RGB }$ as color features works better in one-shot colorization settings, as neural networks seem to learn easily and fast from direct RGB values than from complex color distribution information. In short, our value memory is represented as
\begin{equation}
\begin{aligned}
V={ C }_{ dist }\:\textrm{or}\:{ C }_{ RGB }.
\end{aligned}
\end{equation}
The color information extracted in the above-described manner is later used as a condition given to our colorization networks. Even though either or both of the variants can be used, we will use the notation ${ C }_{ dist }$ for value memory in subsequent equations, so as not to confuse the reader. 

\begin{figure}[t]
\begin{center}
\includegraphics[width=0.9\linewidth]{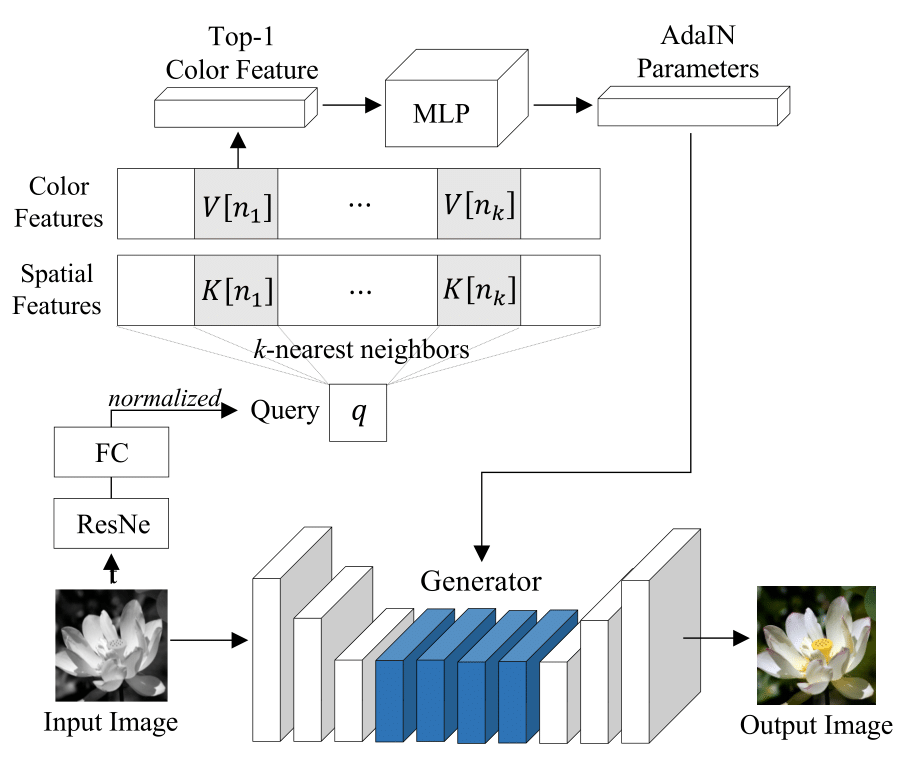}
\end{center}
   \caption{\textbf{How our model works during test time.} The top-1 color feature from our memory is retrieved and given as the condition to our trained generator.}
\label{fig:long}
\label{fig:onecol} \label{fig:test_model}
\end{figure}

\paragraph{Threshold Triplet Loss for Unsupervised Training.}
Previously proposed triplet losses~\cite{schroff2015facenet, kaiser2017learning} aim to make images of the same classes (positive neighbors) closer to each other while making images of different classes (negative neighbors) further away. Likewise, we adopt the triplet loss to maximize similarity between the query and positive key and minimize similarity to the negative key. An existing supervised triplet loss~\cite{kaiser2017learning} introduces the smallest index $p$ where $V[{ n }_{ p }]$ has the same class label as an input query $q$. This would make ${ n }_{ p }$ a positive neighbor of $q$. A negative neighbor of $q$ will be defined as the smallest index $b$ where $V[{ n }_{ b }]$ has a different class label from our query $q$. 

However, this supervised triplet loss requires class label information, leading to its limited applicability in our setting as such information is not available in most data for colorization tasks. For instance, it would be almost impossible to label every single frame of an animation with its class label (i.e., whether a particular character, object, or background appears in a given frame).

To solve this issue, we extend the existing method and propose a threshold-based triplet loss applicable to fully unsupervised settings. Given two images with similar spatial features, we assume that if the distance between their color features are within a certain threshold, then they are more likely to be in the same class than those images with different color distributions. We introduce this threshold as a hyperparameter denoted as $\delta$. As the distance measure ${ C }_{ dist }$ between two color features, we compute the symmetric KL divergence of their color distributions over quantized $ab$ values. For ${ C }_{ RGB }$, we compute color distance using CIEDE2000~\cite{sharma2005ciede2000} by converting RGB values to CIE \emph{Lab} values. In our unsupervised triplet loss setting, we newly define a positive neighbor $n_{p}$ as the memory slot with the smallest index where the distance between $V[{n}_{p}]$ and correct desired value $v$ (i.e., the color feature of the query image) is within a color threshold $\delta$, i.e.,  
\begin{equation}
\begin{aligned}
\mathrm{KL}(V[{ n }_{ p }]\parallel v)<\delta.
\end{aligned}
\end{equation}
Similarly, we define a negative neighbor $n_{b}$ as the memory slot with the smallest index where the distance between $V[{n}_{p}]$ and $v$ exceeds $\delta$, i.e., 
\begin{equation}
\begin{aligned}
\mathrm{KL}(V[{ n }_{ b }]\parallel v)>\delta.
\end{aligned}
\end{equation}
Finally, the threshold triplet loss is defined as
\begin{equation}
\begin{aligned}
{L}_{ t }(q,M,\delta)=\mathrm{max}(q\cdot K[{ n }_{ b }]-q\cdot K[{ n }_{ p }]+\alpha, 0).
\end{aligned}
\end{equation}
This triplet loss minimizes distance between the positive key and the query while maximizing distance between the negative key and the query.

\paragraph{Memory Update.}
Our memory $M$ is updated after a new query $q$ is introduced to the networks. The memory gets updated as follows, depending on whether the color distance between the top-1 value $V[{n}_{1}]$ and the correct value $v$ (i.e., the color feature of the new query image) is within the color threshold.

(i) If the distance between $V[{n}_{1}]$ and $v$ is within the color threshold, 
%we write $(q, v)$ to the memory. 
we update the key by averaging $K[{n}_{1}]$ and $q$ and normalizing it. The age of ${ n }_{1}$ is also reset to zero. In detail, the update when $\mathrm{KL}(V[{ n }_{ 1 }]\parallel{v})< \delta$ is written as
\begin{equation}
\begin{aligned}
K[{ n }_{ 1 }]\leftarrow \frac { q+K[{ n }_{ 1 }] }{ \parallel q+K[{ n }_{ 1 }]\parallel  } ,\quad A[{ n }_{ 1 }]\leftarrow 0.
\end{aligned}
\end{equation}

(ii) If the distance between $V[{n}_{1}]$ and $v$ exceeds the color threshold $\delta$, this indicates that there exists no memory slot that matches $v$ in our current memory. Thus, $(q, v)$ will be newly written in the memory. We randomly choose one of the memory slots with the oldest age (i.e., the least recently used one), denoted as $n_{r}$, and replace that slot with $(q, v)$. We also reset its age to 0. In detail, when $\mathrm{KL}(V[{ n }_{ 1 }]\parallel {v})>\delta$, the update is performed as
\begin{equation}
\begin{aligned}
K[{ n }_{r}]\leftarrow q, V[{ n }_{r}]\leftarrow{ v }_{ q }, A[{ n }_{ r }]\leftarrow0.
\end{aligned}
\end{equation}

%-------------------------------------------------------------------------
\subsection{Colorization Networks}
\paragraph{Objective Function.}
Our colorization networks are conditional generative adversarial networks that consist of a generator $G$ and a discriminator $D$. The discriminator tries to distinguish real images from colored outputs using a grayscale image and a color feature as a condition, while the generator tries to fool the discriminator by producing a realistic colored image given a grayscale input $X$ and a color feature $C$. A smooth $L_{1}$ loss between the generated output $G(x,C)$ and the ground-truth image $y$ is added to the generator's objective function, i.e., 
\begin{equation}
\begin{aligned}
L_{sL1}(y, \hat{y})=\begin{cases}
\frac{1}{2}(y-\hat{y})^{2}\;\;\mathrm{for}\;\;\left|y-\hat{y}\right|\leq\delta\\
\delta\left|y-\hat{y}\right|-\frac{1}{2}\delta^{2}\;\;\mathrm {otherwise}.
\end{cases}
\end{aligned}
\end{equation}
This encourages the generator to produce outputs that do not deviate too far from the ground-truth image. Our full objective function for $D$ and $G$ can be written as
\begin{equation}
\begin{aligned}
{ L }_{ D }=\mathbb{\mathbb{E}}_{ x\sim{P}_{data}}[\textrm{log}D(x,C,y)]\\
+\mathbb{\mathbb{E}}_{x\sim{P}_{data}}[\textrm{log}(1-D(x,C, G(x,C)))],
\end{aligned}
\end{equation}
\begin{equation}
\begin{aligned}
{ L }_{ G }=E_{x\sim{P}_{data}}[\textrm{log}(1-D(x, C, G(x,C) ))] \\ 
+{ L }_{ sL1 }( y, G(x,C) ).
\end{aligned}
\end{equation}
During training, we extract the color feature from the ground-truth image to train $G$ and $D$. During the test time, we utilize the color value retrieved from the memory networks and feed it as the condition to the trained $G$, as shown in Fig.~\ref{fig:test_model}. We adapt the architecture of our generator networks from \cite{isola2017image} and that of the discriminator from \cite{bahng2018coloring}.

% small figure: AdaIn in utilizing color conditions
%\begin{figure}
%\begin{center}
%\includegraphics[width=0.7\linewidth]{cvpr2019AuthorKit2/latex/images/adain_hint.pdf}
%\end{center}
%   \caption{\textbf{Element-wise addition~\cite{zhang2017real} vs. our AdaIN-based method in utilizing color conditions.} Using our AdaIN module to inject color conditions into the grayscale image results in a more vivid and high-quality output. Outputs of both models were obtaiend from the same iterations of each approach.}
%\label{fig:adain_fig}
%\end{figure}

\begin{figure*}
\begin{center}
\includegraphics[width=0.75\linewidth]{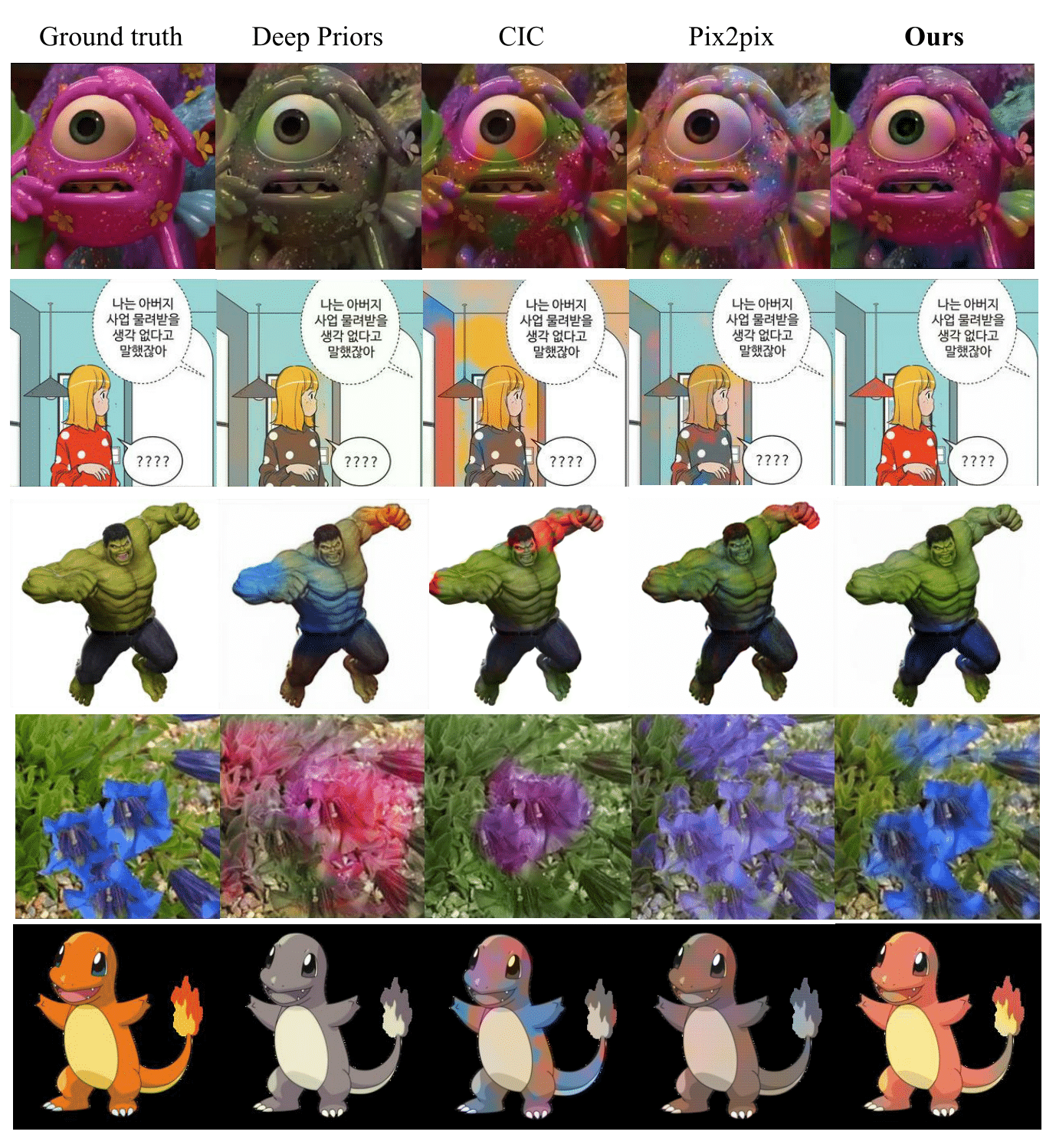}
\end{center}
   \vspace{-0.3em}
   \caption{\textbf{Comparison to baselines on multiple datasets.} We compare our model with other colorization and image translation models: from left, Deep Priors~\cite{zhang2017real}, CIC~\cite{zhang2016colorful}, and Pix2pix~\cite{isola2017image}. Our model particularly excels at capturing and remembering objects that appear only a few times in a training set. The character in the first row is originally green, but he is drenched in pink paint in one scene. Other models color this character in green, but our model \textit{MemoPainter} succeeds in remembering and coloring a scene where he is not green. Our model works even in settings with extremely limited data, where only one data item per class (last row) is available. In this one-shot learning setting, only our model manages to produce vibrant outputs.}
\label{fig:baseline_comparison}
\end{figure*}

\paragraph{Colors as style.} Style transfer is a task of transferring a style of a reference image to a target image. Colorization can be viewed as style transfer, where instead of a particular style, color features are transferred to a target grayscale image. We will regard color as a style and from this perspective, we use AdaIN, which has shown success in style transfer, to effectively inject color information. We compute the affine parameters used in the AdaIN module by directly feeding the color feature to our own parameter-regression networks,  
i.e., 
\begin{equation}
\begin{aligned}
\mathrm{AdaIN}(z,C)=\gamma\left( \frac {z-\mu (z) }{ \sigma (z) }  \right) +\beta,
\end{aligned}
\end{equation}
where $z$ is the activation of the previous convolutional layer, which is first normalized. Then it is scaled by $\gamma$ and shifted by $\beta$, which are parameters generated by a multilayer perceptron adapted from \cite{huang2018multimodal}. Compared to existing colorization models~\cite{bahng2018coloring, zhang2017real} that incorporate color conditions via a simple element-wise addition, AdaIN allows the model to produce vivid colorizations as shown in Fig.~\ref{fig:baseline_comparison}.

\begin{figure}
\begin{center}
\includegraphics[width=1.0\linewidth]{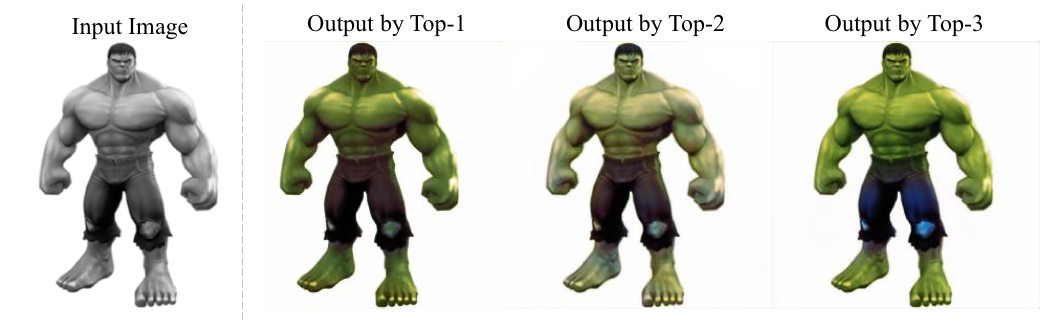}
\end{center}
   \vspace{-0.3em}
   \caption{\textbf{Colorization results using the top-3 memory slots.} The memory networks can retrieve appropriate color features for a given input. Different memory slots may be used to produce diverse results. All other samples in the paper are colored using the top-1 memory slot.}
\label{fig:color_top3}
\end{figure}

% long figure
% 꽃 사진은 rare 한 class 인거 강조하기
%\thispagestyle{empty}

%-------------------------------------------------------------------------
\begin{figure*}
\begin{center}
\includegraphics[width=1.0\linewidth]{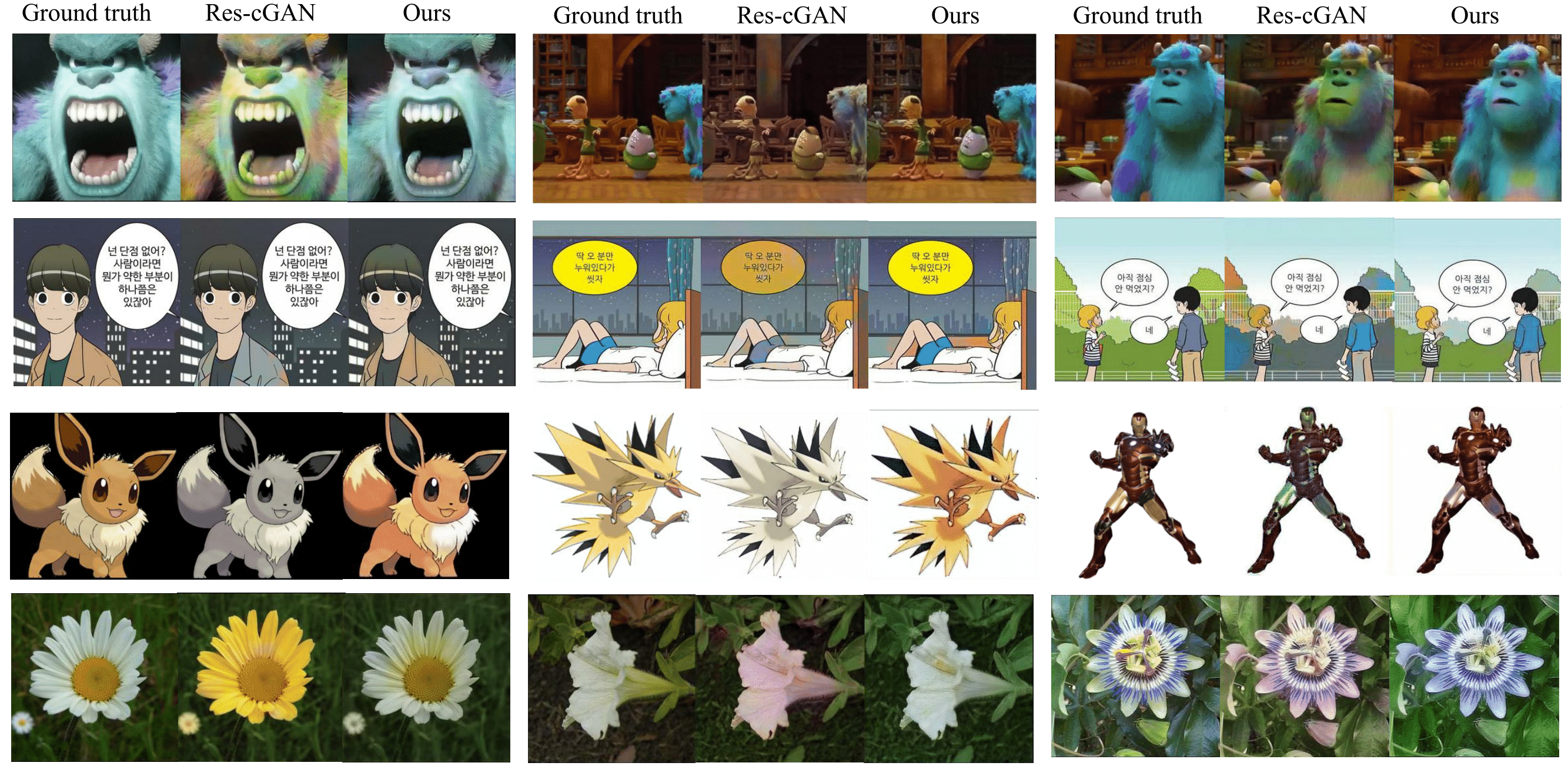}
\end{center}
   \caption{\textbf{Analysis on memory networks.} We apply our model to a wide variety of datasets to show its applicability to different types of images ranging from diverse cartoons to real-world images. Our model shows superior performance especially in few-shot settings (first and third rows) when compared to our colorization networks (Res-cGAN) without memory networks. Colorization networks find it difficult to produce outputs with vivid color.}
\label{fig:short} \label{fig:memory_analysis}
\end{figure*}

\section{Experiments} \label{sec:experiments}
Our experiments consist of an ablation study on memory networks, analysis on the threshold triplet loss, and both quantitative and qualitative comparisons on three baseline models.

\subsection{Qualitative Evaluation}
\subsubsection{Datasets}
\vspace{-0.5em}
We perform experiments on five different datasets and compare our model performance on diverse settings (abundant data, few-shot, and one-shot).

\vspace{1em}
\noindent
\textbf{Oxford102 Flower Dataset.} The Flower dataset~\cite{Nilsback08} consists of 102 flower classes. Each class has 40 to 258 images. The class labels are not used in our experiments.

\vspace{1em}
\noindent
\textbf{Monster Dataset.} 1,315 images are collected from the trailer of the movie \textit{Monsters, Inc.}~\cite{monster} to perform colorization on animations. An image frame is extracted every two seconds to reduce excessive redundancy in the dataset.

\vspace{1em}
\noindent
\textbf{Yumi Dataset.} We collect 9,955 images of the cartoon \textit{Yumi's Cells}~\footnote{https://comic.naver.com/webtoon/list.nhn?titleId=651673} to perform colorization on cartoons. It consists of images from 329 episodes, and each image is a single frame from of a sequence of each episode.

\vspace{1em}
\noindent
\textbf{Superheroes Dataset.} We collect images of superhero characters to perform few-shot colorization. It consists superhero images from seven categories with less than five images per category.

\vspace{1em}
\noindent
\textbf{Pokemon Dataset.} We utilize the Pokemon dataset~\footnote{https://www.kaggle.com/kvpratama/pokemon-images-dataset} for one-shot colorization, which consists of 819 classes with a single image per class. Additional images are crawled from the internet to construct the test set. 

\vspace{1em}
\subsubsection{Analysis on Memory Networks}
\vspace{0.0001cm}
We run an ablation study to analyze the effect of augmenting colorization networks with memory. As shown in Fig.~\ref{fig:memory_analysis}, we compare our proposed model \textit{MemoPainter} against our colorization networks (Res-CGAN) without memory augmentation. Our memory-augmented networks are able to produce superior results on a wide variety of datasets from diverse cartoons to real-world images. In particular, it can accurately color an image even with only a single or few instances. Even though Res-cGAN produces high-quality colorization in most cases, it fails to preserve the ground-truth color of rare instances or completely fails in one-shot learning settings (e.g., results on the Pokemon dataset).

\begin{figure}
\begin{center}
\includegraphics[width=1.0\linewidth]{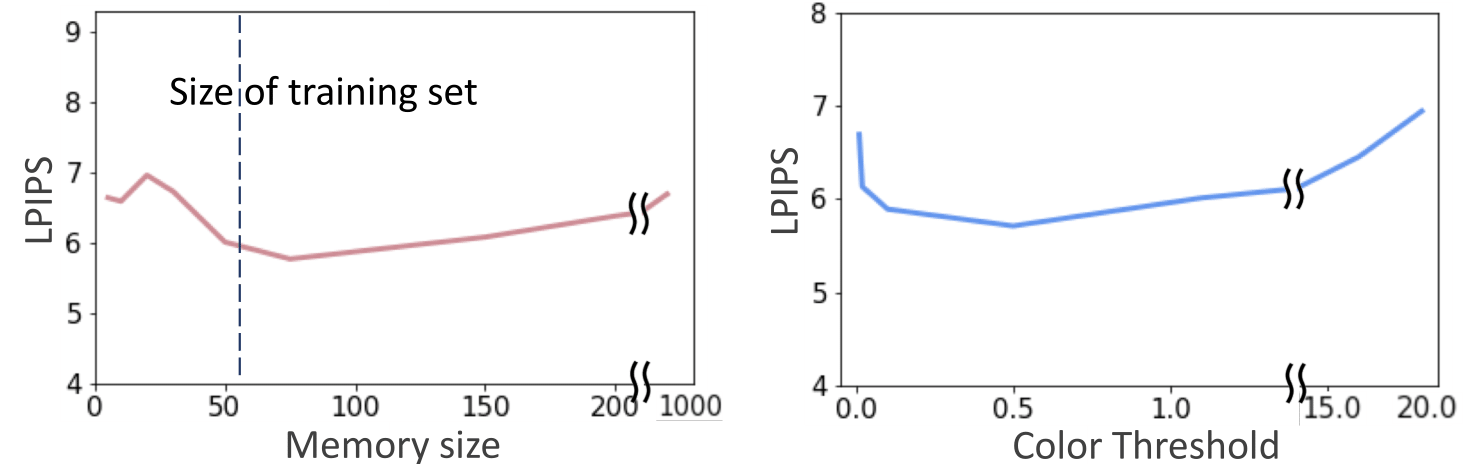} \rule{0.9\linewidth}{0pt}
\end{center}
   \vspace*{-0.4cm}
   \caption{\textbf{Analysis of memory size and color threshold.} LPIPS scores are similar across various hyperparameters of the memory networks. Quality drops (high LPIPS) only with excessively small or large hyperparameters.}
\label{fig:hyper}
\end{figure}

Moreover, an analysis on two hyperparameters (memory size and color threshold) is shown in Fig.~\ref{fig:hyper}. Performance is measured by comparing average Learned Perceptual Image Patch Similarity(LPIPS)~\cite{zhang2018unreasonable}. Results show that LPIPS scores are stable across a wide range of hyperparameters and the model does not overfit to a particular color threshold or memory size. 
 
\begin{figure}[t]
\begin{center}
\includegraphics[width=0.75\linewidth]{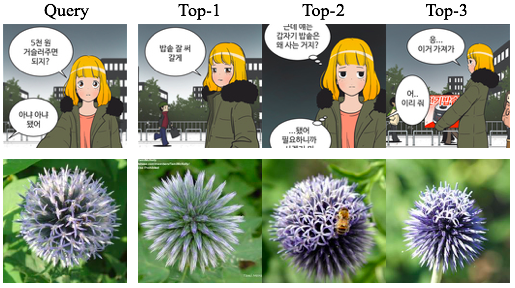}
\end{center}
   \caption{\textbf{Validation of our assumption on threshold triplet loss.} We demonstrate the corresponding images of the top-3 color features retrieved from our memory networks. By using the threshold triplet loss, our memory networks are trained to retrieve color features highly relevant to the content of the query image.}
\label{fig:long}
\label{fig:onecol} \label{fig:threshold_results}
\end{figure}
% top 1 mem slot 뿐만 아니라 2,3 slot 들도 비슷한 색을 뽑아내는 것을 보여준다.

\subsubsection{Analysis on Threshold Triplet Loss (TTL)}
The assumption behind the threshold triplet loss is that those images with (i) similar spatial features (i.e., the $k$ nearest neighbors) and (ii) similar color features (i.e., color distance within a particular threshold) are likely to be in the same class. To confirm our assumption, we demonstrate the corresponding images of the top-3 color features retrieved from our memory networks during training. In this setting, we do not update the keys in the case $\mathrm{KL}(V[{ n }_{ 1 }]\parallel{v})<\delta$, so as to show its corresponding image. As shown in Fig.~\ref{fig:threshold_results}, we can see that the corresponding images of the top-3 color features have the same class as the query image. In particular, examples in the first row show that the top-3 images share the same character as well as similar clothes, objects, and backgrounds. This shows that TTL allows our memory networks to retrieve color features relevant to the content of the query image, being able to color rare instances even if it was presented just once in the training data. 

We also quantitatively validate the assumption of TTL by measuring the classification accuracy, evaluating whether the class of the images corresponding to the top-1 memory slot and the ground-truth label of the query are identical. During training, we additionally save the query's class to compute classification accuracy. At test time, we compute the accuracy by computing the percentage of queries that has the same class as the top-1 memory slot. As an upper bound of our~\textit{unsupervised} method, we use a~\textit{supervised} version of our model. This version stores class values in the value memory and updates memory as presented by the life-long memory module made for few-shot classification~\cite{kaiser2017learning}. Although our model is not specifically made for classification tasks, Table~\ref{table:triplet} shows that our unsupervised model (first row) retrieves the memory with the same class as accurately as the supervised method (second row) across a different number of classes and training sets.
Note that TTL works on datasets that have a consistent color for every classes, such as animation characters and flower species. TTL would not work on datasets where the colors are arbitrary. 

\subsubsection{Qualitative Comparisons}
We qualitatively compare \textit{MemoPainter} with three baselines: Deep Priors~\cite{zhang2017real}, CIC~\cite{zhang2016colorful}, and Pix2pix~\cite{isola2017image}. Fig.~\ref{fig:baseline_comparison} shows qualitative results on multiple datasets. It shows that our model particularly excels at coloring rare instances in a training set. The Result from the monster dataset (first row) shows a rare scene where the main character (originally green) is drenched in pink paint. Although other models color this character in green, \textit{MemoPainter} is able to remember this rare instance and color the character properly. Similarly, results from both the second and the fourth rows show that our model is able to remember rare classes along with minor details (i.e., even the clothes, objects, and backgrounds in a cartoon frame) while every other baseline fails to do so. Furthermore, our model is capable of producing high-quality results in both few-shot and one-shot learning settings compared to existing methods. Our model successfully produces accurate colorization given extremely limited data, e.g., given less than five training images per class (third row) or only a single data item per class (last row). In both cases, \textit{MemoPainter} is the only model that can consistently produce accurate and vibrant colorization results.
%%%%%%%%%%%%%%%%%%%%%%%%
%% User study & LPIPS %%
%%%%%%%%%%%%%%%%%%%%%%%%

\begin{table}[]
\centering
\begin{tabular}{ccccc}
\hline
\multicolumn{1}{l}{} & \multicolumn{2}{c}{\textbf{One-shot}} & \multicolumn{2}{c}{\textbf{Few-shot}} \\ \hline
                     & User-study        & LPIPS             & User-study        & LPIPS             \\ \hline
Ours                 & \textbf{75\%}     & \textbf{8.48}     & \textbf{71\%}     & \textbf{1.34}     \\
CIC                  & 10\%              & 9.89              & 7\%               & 1.80            \\
Pix2pix              & 5\%               & 13.47             & 16\%              & 2.34               \\
Deep Prior           & 10\%              & 19.26             & 4\%               & 2.03              \\ \hline
\end{tabular}
    \vspace*{+0.3cm}
    \caption{\textbf{Quantitative comparisons with the state-of-the-art.}
    User study (higher is better) and LPIPS perceptual distance metric~\cite{zhang2018unreasonable} (lower is better) shows superiority of our method.}
\label{table:quant}
\end{table}

\begin{table}[]
\centering
\begin{tabular}{ccccc}
\hline
              & \multicolumn{2}{c}{\textbf{5-way}} & \multicolumn{2}{c}{\textbf{15-way}} \\ \hline
              & 5-shot      & 10-shot     & 5-shot       & 10-shot     \\ \hline
Ours (Unsup.) & 87.50\%     & 87.50\%     & 69.44\%      & 70.83\%     \\ 
Ours (Sup.)   & 91.66\%     & 87.50\%     & 72.22\%      & 75.00\%     \\ \hline
\end{tabular}
    \vspace*{+0.001cm}
    \caption{\textbf{Classification accuracy of the threshold triplet loss.}}
\label{table:triplet}
\vspace*{-0.4cm}
\end{table}

\subsubsection{Quantitative Evaluation}
To quantitatively evaluate colorization quality, we conduct a user study with 30 participants, each answering 40 questions. We give a random source image and its corresponding colored outputs from our model and baselines. We then ask which generated output has the highest quality while maintaining the color identity of the source image (e.g., Hulk is green). We also compare the LPIPS distance~\cite{zhang2018unreasonable} which is closer to human perception unlike MSE-based metrics. We compute the average LPIPS between the input image and its corresponding colored image. Table~\ref{table:quant} shows that our model is superior to the state-of-the-art across both measures. 

%-------------------------------------------------------------------------
\section{Conclusions}
Results of this paper suggest that colorization networks with memory networks are a promising approach for practical applications of colorization models. We stress the importance of colorization models working with little data so that they can be used in coloring animations and cartoons. \textit{MemoPainter} works on a wide variety of images, thus bearing great potentials in various applications that require few-shot colorization. 
\vspace{-0.7em}
\paragraph{Acknowledgements.} This work was partially supported by the National Research Foundation of Korea (NRF) grant funded by the Korean government (MSIP) (No. NRF2016R1C1B2015924). We thank all researchers at NAVER WEBTOON Corp., especially Sungmin Kang. Jaegul Choo is the corresponding author. 
%-------------------------------------------------------------------------

% {\small
% \bibliographystyle{ieee}
% \bibliography{egbib}
% }

\end{document}